\documentclass[10pt,twocolumn,letterpaper]{article}

\usepackage{cvpr}              

\usepackage[accsupp]{axessibility}  
\usepackage{bm}
\usepackage{amssymb}
\usepackage{makecell}
\usepackage{adjustbox}
\usepackage{booktabs} 
\usepackage{multirow} 
\usepackage{subcaption}
\usepackage{comment}

\definecolor{cvprblue}{rgb}{0.21,0.49,0.74}
\usepackage[pagebackref,breaklinks,colorlinks,allcolors=cvprblue]{hyperref}

\def\paperID{13200} 
\def\confName{CVPR}
\def\confYear{2025}

\title{Towards Human-Understandable Multi-Dimensional Concept Discovery}

\author{Arne Grobrügge\textsuperscript{1}
\and
Niklas Kühl\textsuperscript{2}
\and
Gerhard Satzger\textsuperscript{1}
\and
Philipp Spitzer\textsuperscript{1}\thanks{{\tt\small Correspondence: philipp.spitzer@kit.edu}}
\\
\and
\textsuperscript{1}Karlsruhe Institute of Technology, Germany
\and 
\textsuperscript{2}University of Bayreuth, Germany
}

\begin{document}
\maketitle
\begin{abstract}
Concept-based eXplainable AI (C-XAI) aims to overcome the limitations of traditional saliency maps by converting pixels into human-understandable concepts that are consistent across an entire dataset. A crucial aspect of C-XAI is completeness, which measures how well a set of concepts explains a model's decisions. Among C-XAI methods, Multi-Dimensional Concept Discovery (MCD) effectively improves completeness by breaking down the CNN latent space into distinct and interpretable concept subspaces. However, MCD's explanations can be difficult for humans to understand, raising concerns about their practical utility. To address this, we propose Human-Understandable Multi-dimensional Concept Discovery (HU-MCD). HU-MCD uses the Segment Anything Model for concept identification and implements a CNN-specific input masking technique to reduce noise introduced by traditional masking methods. These changes to MCD, paired with the completeness relation, enable HU-MCD to enhance concept understandability while maintaining explanation faithfulness. Our experiments, including human subject studies, show that HU-MCD provides more precise and reliable explanations than existing C-XAI methods. The code is available at \url{https://github.com/grobruegge/hu-mcd}.

\end{abstract}    
\section{Introduction}
\label{sec:intro}


In recent years, industry and research have witnessed an exponential growth of Machine Learning (ML), particularly accelerated by the advancements of Deep Learning (DL). ML models have proven to solve various problems given sufficient data.
However, these models are not failure-free.
Previous work has identified various failure cases, including vulnerability to adversarial attacks~\citep{szegedy2013intriguing} or biased data~\citep{buolamwini2018gender} and reliance on spurious features~\citep{xiao2021noise} that leads to shortcut learning~\citep{geirhos2020shortcut}.
In high-stake sectors, such as autonomous driving and healthcare, failures could lead to catastrophic outcomes \citep{morrison2024impact, spitzer2024don}.
To prevent such risks, ensuring reliable transparency of ML models is essential. 
With the recent European regulations---including the General Data Protection Regulation (GDPR)~\citep{kaminski2021right} and the European AI act~\citep{kop2021eu}---this need for transparency has made eXplainable Artificial Intelligence (XAI) a central topic in ML research~\citep{tjoa2020survey}. XAI methods aim to provide transparency of ML models by providing insights into their decision-making processes.

\noindent
Due to their convenience, \emph{post-hoc} explanations that do not require modifying the underlying model architecture are widely used in the computer vision domain, particularly \emph{local} XAI methods that provide explanations for single predictions~\citep{adadi2018peeking, zhang2021survey, spitzer2024transferring}. For instance, saliency maps highlight areas in images that are significant for the model's predictions. However, there is a growing consensus that these methods do not provide understandable explanations. \citet{adebayo2018sanity} find that these maps are independent of both the model and the data-generating process, questioning the reliability of such approaches. Moreover, local explanation methods are vulnerable to human confirmation bias~\citep{vielhaben2023multi}---the tendency of humans to favor information that confirms their preexisting assumptions while disregarding contradictory evidence.
Finally, the human interpretation of local explanations, such as attribution maps on individual instances, poses challenges and may lead humans to draw contradictory conclusions~\citep{kim2018interpretability, hase2020evaluating, kim2022hive}.
Highlighting the location of important regions within an image---\textit{where} the model ``looks''---is thus not sufficient for humans to interpret the reasoning of a model. Humans also require the semantic content---\textit{what} the model ``sees''~\citep{fel2024holistic}.

\noindent
To address the shortcomings of local methods, Concept-based XAI (C-XAI) has emerged as a promising line of research within the area of \emph{global post-hoc} explanations---which take a more holistic approach and explain the overall decision logic of ML models~\citep{schwalbe2023comprehensive}. \textit{Concepts} refer to patterns learned by the model that can be associated with high-level and human-understandable visual attributes. For instance, in the medical field, the recognition of such patterns is crucial to assist clinicians in improving diagnostic accuracy. \citet{lucieri2022exaid} utilize C-XAI in a medical scenario and demonstrate its use in dermatology.  While early methods used pre-defined concept datasets~\citep{kim2018interpretability, bau2017network, zhou2018interpretable} against which the model is evaluated, more recent work has developed frameworks for automatic concept discovery \citep{ghorbani2019towards, zhang2021invertible, fel2023craft, vielhaben2023multi}. First, these frameworks identify visual attributes in images of specific classes for a given task. Then, they cluster similar attributes to form meaningful concepts related to that task. However, they have a conflicting relationship between discovering human-\textbf{understandable} concepts and \textbf{faithfully} quantifying their significance to model predictions, often prioritizing one over the other. 

\noindent 
For instance, Automatic Concept-based Explanations (ACE)~\citep{ghorbani2019towards} uses image segmentation clustering for concept discovery. While obtaining promising results in terms of understandability, the segments must be inpainted and rescaled to meet the model input requirements, resulting in noise that distorts the model's predictions. Furthermore, to ensure understandable concepts, ACE uses several heuristics to exclude outliers but does not consider the degree of information loss, thus raising concerns regarding the faithfulness of its explanations. More recent work has addressed some of the limitations of ACE. In particular, Invertible Concept-based Explanations (ICE)~\citep{zhang2021invertible} replaces image segments with hidden feature maps, and Concept Recursive Activation FacTorization for Explainability (CRAFT)~\citep{fel2023craft} utilizes quadratic image patches to circumvent inpainting requirements. Nonetheless, a key challenge remains to guarantee faithful explanations: the quantification of the \textbf{completeness} of a concept set, \ie, the extent to which these concepts are sufficient to explain the model's predictions. Multi-Dimensional Concept Discovery (MCD)~\citep{vielhaben2023multi} generalizes upon ICE and incorporates a completeness relation, highlighting the superior faithfulness of their concepts compared to previous methods. However, MCD does not quantitatively assess the understandability of its discovered concepts.



\noindent
In this work, we propose a novel framework that provides both understandable and faithful explanations: Human-Understandable MCD (HU-MCD). 
To discover concepts that are human-understandable, we use the Segment Anything Model (SAM)~\citep{kirillov2023segment}.
To overcome the noise introduced by the use of rescaled or inpainted images, we employ a novel input masking scheme tailored for Convolutional Neural Networks (CNNs)~\citep{balasubramanian2023towards}. Subsequently, we adopt the MCD framework, which enables both local and global concept importance scoring to quantify the significance of each concept for the model's predictions. Unlike ACE, HU-MCD incorporates a \textit{completeness} relation, allowing to account for potential information loss during concept discovery, further enhancing the explanations' faithfulness.


\noindent
We evaluate HU-MCD on the ImageNet1k~\citep{deng2009imagenet} dataset and demonstrate that HU-MCD outperforms state-of-the-art methods in both the understandability of discovered concepts and the faithfulness in attributing their importance to the model. 
Furthermore, we benchmark HU-MCD using Concept Deletion (C-Deletion) and Concept Insertion (C-Insertion), thereby demonstrating that the concept importance scores faithfully represent the model's reasoning. 

\noindent
Overall, our main contributions are threefold: (1) We introduce HU-MCD---a framework for automatic completeness-aware concept-based explanations that uses SAM for human-understandable concept discovery. By using SAM, the manual labeling effort in real-world settings can be reduced. (2) To ensure explanations that faithfully relate to the model's decision-making, we use an input masking scheme tailored for CNNs that effectively mitigates noise introduced by the segmentation masks. (3) We design a human subject study and conduct extensive experiments on established benchmarks to verify the understandability and faithfulness of the concepts generated by HU-MCD. Thereby, HU-MCD takes further steps towards aligning AI decisions with legal regulations (e.g., EU AI Act).
\section{Related Work}
\label{sec:rel_work}

\textbf{Supervised Post-Hoc Concept Analysis.} Recent research demonstrated that CNNs can encapsulate human-understandable concepts without being explicitly trained on these concepts~\citep{zhou2014object}. This discovery led to the development of several XAI methods, aiming to discover these concepts and measure their influence on model predictions~\citep{bau2017network, fong2018net2vec, zhou2018interpretable, kim2018interpretability}. Notably, the Testing with Concept Activation Vectors (TCAV) framework~\citep{kim2018interpretability} introduces the notion of Concept Activation Vectors (CAVs)---the weights of a linear classifier used to separate activations corresponding to a specific concept from those corresponding to random data within the activation maps of a neural network's final convolutional layer.
CAVs provide a formalized method for representing and quantifying concepts, enabling the interpretation of model behavior in terms of human-understandable features.

\noindent
\textbf{Unsupervised Post-Hoc Concept Discovery.} While the methods above rely on the availability of a human-defined concepts dataset, subsequent work aimed to eliminate this dependency by automatically discovering concepts. ACE~\citep{ghorbani2019towards} uses superpixel segmentation of class images, clusters their embeddings, and groups similar segments as examples of a concept, which are then analyzed using TCAV. However, the clustering step requires that images are cropped, mean-padded, and resized to the model's input size. These image manipulations distort the aspect ratio, introduce noise, and discard the overall scale ratio. Additionally, ACE applies several heuristics to discard irrelevant segments and clusters but does not account for the information loss during this process. More recent work addresses these shortcomings~\citep{yeh2020completeness,zhang2021invertible,fel2023craft,vielhaben2023multi}. \citet{yeh2020completeness} builds on ACE by introducing the notion of \textit{completeness}---the extent to which concept scores serve as sufficient statistics for recovering the model's prediction. However, they provide limited qualitative results and lack a rigorous human-subject study comparing their approach to similar work.
ICE~\citep{zhang2021invertible} applies Non-negative Matrix Factorization (NMF) on feature maps to identify concepts by disentangling frequently appearing directions within the feature space. CRAFT~\citep{fel2023craft} combines ACE and ICE by applying NMF to feature vectors of image sub-regions, thereby eliminating the necessity for a baseline value to inpaint masked regions for image segments but still requiring rescaling. Unlike ICE, CRAFT uses Sobol indices instead of TCAV but lacks a completeness relation. Instead of representing concepts as a single direction in the feature space, MCD~\citep{vielhaben2023multi} allows concepts to lie on a hyperplane spanned across different convolutional channel directions, thus generalizing ICE. This is realized by Sparse Subspace Clustering (SSC) for feature vector clustering and a subsequent Principle Component Analysis (PCA) for cluster basis derivation. Observing the projection into the subspace not covered by the concepts allows for defining a global completeness score directly on the model's parameters, which differs from the original completeness definition~\citep{yeh2020completeness}. Similar to ICE, MCD aims to mitigate the noise introduced by rescaling and inpainting image segments. However, their methodology lacks an evaluation of the understandability of the concept. Finally, Segment Any Concept (SAC)~\citep{sun2024explain} provides \textit{local} post-hoc concept explanation. Their definition of concepts differs from the methodologies discussed earlier, focusing exclusively on individual image regions within single images rather than on shared patterns across multiple instances. 

\noindent
\textbf{Self-Interpretable Concept Models.} All the methods above make no modification to the underlying model architecture. An alternative approach is to re-design the architecture such that the decision process is inherently linked to concepts' representations. For instance, Concept Bottleneck Models (CBM)~\citep{koh2020concept} introduce a concept bottleneck layer, where single neurons are explicitly linked to pre-defined concepts which has inspired several subsequent studies~\citep{espinosa2022concept, yuksekgonul2022post}. While CBMs require concept annotation for the training dataset, self-interpretable models can also be designed in an unsupervised manner~\citep{chen2019looks, yang2023language, oikarinen2023labelfree}.
\section{Proposed Method}
\label{sec:proposed_method}

\begin{figure*}[t]
    \centering
    \includegraphics[width=0.85\textwidth]{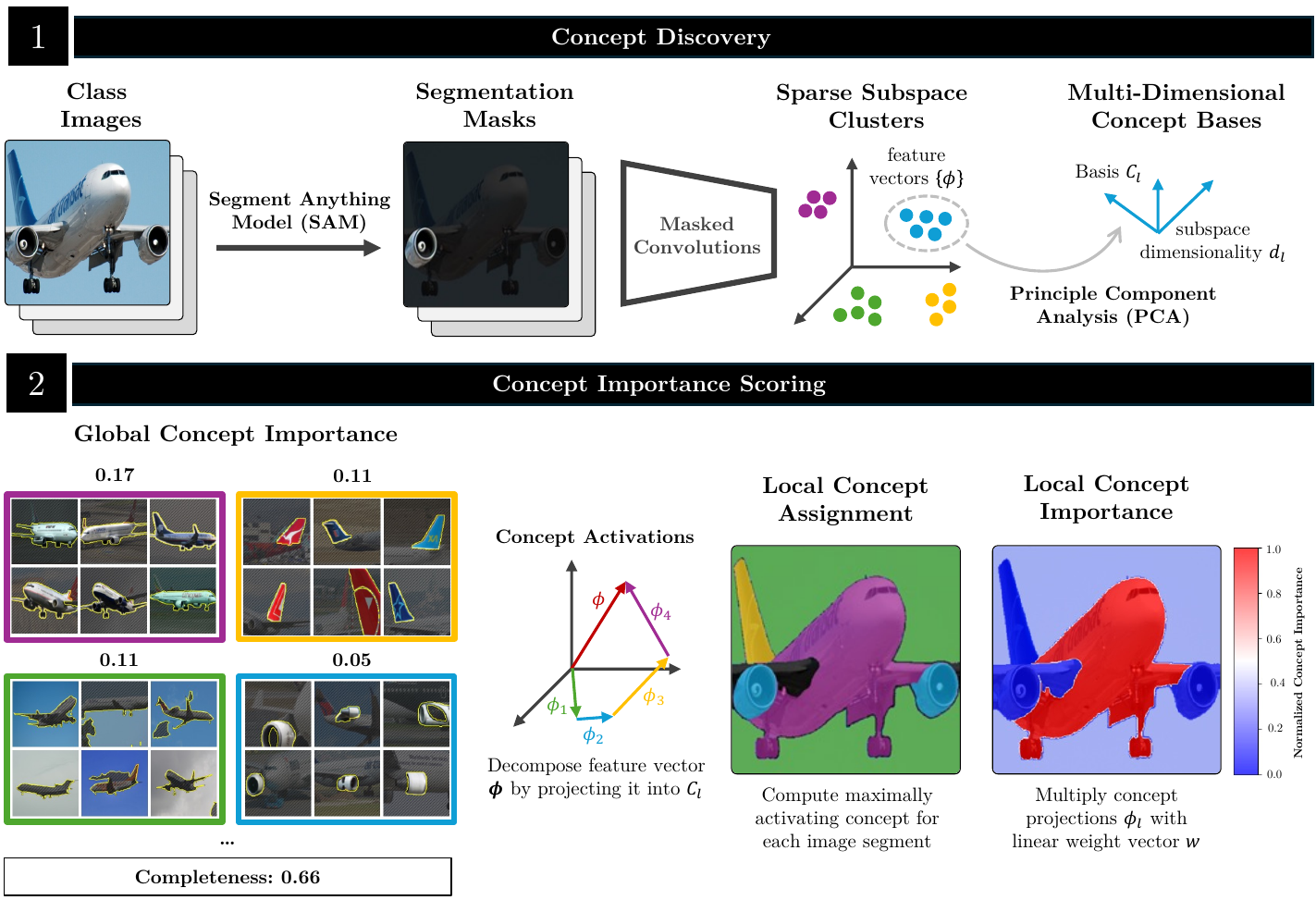}
    \caption{Overview of HU-MCD.}
    \label{fig:method_overview}
\end{figure*}

Building on the objectives of discovering human-\textit{understandable} concepts and \textit{faithfully} attributing their contribution to the model's prediction strategy, we propose Human-Understandable Multi-Dimensional Concept Discovery (HU-MCD). As shown in
Figure~\ref{fig:method_overview}, we distinguish two stages for HU-MCD to align
with existing literature~\citep{fel2024holistic}. First, we discover relevant concepts by segmenting class images using SAM~\citep{kirillov2023segment}, a foundation model for instance segmentation. We cluster them using SSC based on their representation within the feature space of the model. To avoid introducing noise that might distort the model's prediction by rescaling and inpainting image segments, we employ an input masking scheme specifically designed for CNNs~\citep{balasubramanian2023towards}. Second, we adapt the MCD framework proposed by~\citet{vielhaben2023multi}, which allows for both local and global concept importance scoring and incorporates a completeness relation by decomposing the model’s feature space into multi-dimensional concepts.

\subsection{Concept Discovery} 

We choose a set of images that encapsulate the concepts against which the model will be tested. The selection of samples is not constrained, giving users the flexibility to choose class-specific samples or utilize the entire training set to derive class-agnostic concepts. Inspired by ACE, we use a segmentation algorithm to obtain a dataset comprising distinct image regions acting as concept candidates. In particular, we use SAM~\citep{kirillov2023segment}, as it provides precise and comprehensive instance segmentation with strong zero-shot generalization demonstrated across a broad spectrum of tasks~\citep{sun2024explain}. Using the masks provided by SAM, we select, for each image, the most granular decomposition, considering all masks covering at least $1\%$ of the image.

\noindent
Concepts are inherently linked to the hidden representation of intermediate feature layers. Previous work has demonstrated that state-of-the-art CNNs learn to represent different features of the data by mapping distinct concepts to different regions of the embedding space~\citep{zhang2018unreasonable}. Thus, we employ SSC to ensure that the data points within different clusters lie on a union of distinct low-dimensional subspaces embedded within a higher-dimensional space~\citep{vielhaben2023multi}, effectively grouping perceptually similar segments as entities of the same concept. Instead of opting for an arbitrary cluster count (as seen in prior studies such as 10 in \citet{zhang2021invertible} or 25 in \citet{ghorbani2019towards} and \citet{fel2023craft}), we leverage the robust segmentation capabilities of SAM and determine the number of clusters based on the average number of segments per image.\looseness=-1

\noindent
Processing image segments presents the challenge of passing irregularly shaped regions through CNN models while extracting feature embeddings. This task is complex due to the fixed-size input requirement of CNN architectures, necessitating rescaling and/or the incorporation of baseline colors to fill masked-out regions, as implemented in ACE. However, many baseline colors are not truly neutral~\citep{jain2022missingness}, potentially introducing artifacts that can bias the model’s predictions. Approaches to address this problem include classical imputation algorithms~\citep{bertalmio2001navier} as suggested by~\citet{vielhaben2023multi} or deep generative models~\citep{chang2018explaining}. While these can be effective, they may inadvertently reveal hidden information by recovering masked-out regions or require expensive model training.

\noindent
To improve upon ACE, we aim to avoid introducing spurious cues from the mask shape or color. Recent studies indicate that Vision Transformers are less susceptible to masking patterns and colors~\citep{jain2022missingness}, partially because they can omit patch tokens. In a similar spirit, \citet{balasubramanian2023towards} propose a \textit{layer masking} scheme for CNNs: rather than inserting a baseline color, both the input image and an accompanying mask are propagated through each layer. This effectively simulates running the CNN on an irregularly shaped input---ignoring any activations arising purely from masked regions. By capitalizing on the hierarchical nature of CNNs (whose stacked convolutional layers expand their receptive fields gradually), the masking scheme retains only those values dependent on the unmasked input while discarding those reliant solely on the masked-out areas. Empirically, this leads to better preservation of model accuracy and more \textit{faithful} explanations.

\noindent
A key challenge in this setup are convolutions near mask edges. Discarding edge convolutions can cause the unmasked portion to rapidly diminish, whereas propagating edges may introduce artifacts by inadvertently revealing masked regions to the model. To mitigate such effects, \citet{balasubramanian2023towards} propose \textit{neighborhood padding}, where the unmasked boundary pixels are padded with an average of their adjacent (non-masked) neighbors. This padding is iterated until it reaches the convolution kernel size. However, given that SAM often produces masks closely aligned with true object edges, extensive padding would remove relevant shape information. For example, an accurately segmented mask of a car tire---once padded to the kernel size---could yield a uniform fill, discarding shape detail entirely.
We thus propagate the \textit{un}masked image along with the mask to the first convolutional layer and only apply the masking scheme thereafter. For instance, in a ResNet50 model (with a $7\times7$ kernel in the first convolution), this grants the kernel access to a narrow band of context around the mask, preventing over-aggressive removal of relevant shape cues. If a particular mask covers more than $25\%$ of the image, we shrink it by the convolution kernel size to avoid exposing large portions of an object’s outline when only the background was intended to be masked. Although not flawless, this pragmatic strategy addresses most typical scenarios: small objects remain intact, and large background regions avoid inadvertently revealing the object shape.

\subsection{Concept Importance Scoring}

To quantify how identified concepts influence the predictions of a model, we adapt the MCD framework. Initially, concepts represent collections of segments associated with similar visual patterns. To achieve independence from individual segments, MCD employs PCA on the hidden representations of cluster members, retaining the top principal components as a representative subspace basis, capturing recurring activation patterns. Repeating this process across all clusters yields a set of subspaces that collectively form a comprehensive basis for the feature space. An additional orthogonal complement subspace captures residual information not represented by the identified concepts, ensuring a complete representation.
Building upon this decomposition, \citet{vielhaben2023multi} introduced two complementary metrics: \emph{concept activation}, which quantifies a concept's presence, and \emph{concept relevance}, which assesses the significance of a concept in predicting class labels.

\noindent
In the original MCD implementation, no image segmentation is utilized; instead, an entire image is processed through the network to generate a feature map, treating each spatial location as a separate feature vector. MCD then decomposes these vectors to compute concept activation and relevance scores, resulting in grid-aligned heatmaps upscaled for visualization \citep{selvaraju2017grad}. This approach, however, tends to yield ambiguous and block-like regions, as the grid alignment may not correspond to actual object boundaries or meaningful parts.
In contrast, our proposed method integrates SAM to guide the discovery of semantically meaningful image regions, thus addressing these interpretability limitations (see Section~\ref{sec:eval}). Consequently, we adapted the original metrics to operate explicitly on image segments, as detailed below.

\noindent
\textbf{Concept Activation.}
Concept activation measures the presence of a concept within a given image segment. Specifically, each concept corresponds to a distinct low-dimensional subspace within the network's hidden layer. By projecting an image segment’s hidden representation onto this subspace, we quantify how strongly a concept is expressed. Applying this procedure across all segments generates a concept activation map highlighting regions where each concept is predominantly active. By identifying the concept with the maximum activation score for each segment, we split images into distinct concept regions. Concept prototypes---segments exhibiting the highest activation scores within a sample set---provide intuitive visualizations.

\noindent
\textbf{Local Concept Relevance.}
While concept activation indicates the \emph{presence} of a concept, it does not directly reveal the concept’s influence on the model’s prediction. To address this,~\citet{vielhaben2023multi} decompose the final hidden layer representation---followed only by a linear mapping to scalar class scores (\emph{logits})---into contributions from each concept subspace. This decomposition generates \emph{local concept relevance} scores, which quantify each concept’s impact on the classification decision at the instance level. Importantly, summing these relevance contributions precisely reconstructs the full logit value, satisfying a completeness criterion. Thus, local relevance scores measure how individual concepts contribute positively or negatively toward a classifier’s decision. Applying this analysis segment-wise results in concept relevance heatmaps.

\noindent
\textbf{Global Concept Relevance.}
In contrast to local relevance, \emph{global} concept relevance quantifies concept importance at the class level by projecting the final classification layer’s weight vector onto the respective concept subspaces. Similar to local relevance, summing global relevance scores across concepts fully reconstructs the classifier’s predictive capability, fulfilling a global completeness criterion based on the model's parameters.

\noindent
A key advantage of this decomposition framework is its inherent \emph{completeness relation}, ensuring that summing either local or global concept relevance values reproduces the original model outputs (logits or weights). Thus, HU-MCD combines \emph{human-interpretability}---by leveraging SAM-generated fine-grained, semantically meaningful segments---and \emph{faithful} interpretation---through adopting MCD’s completeness properties. Furthermore, our input masking strategy supports model fidelity by minimizing noise from baseline colors or mask boundaries.
\section{Evaluation}
\label{sec:eval}

We evaluate HU-MCD's (1) \textit{understandability} of the automatically discovered concepts from a human perspective and (2) \textit{faithfulness} of the concept importance scores in explaining the model prediction. We run our experiments on the ImageNet1k dataset~\citep{deng2009imagenet} using a selection of ten classes that roughly align with CIFAR10 classes\footnote{airliner, beach wagon, hummingbird, Siamese cat, ox, golden retriever, tailed frog, zebra, container ship, police van} as proposed by~\citet{vielhaben2023multi}. As model architecture, HU-MCD uses ResNet50 with the weights provided by the Python library \textit{timm}~\citep{rw2019timm}. For segmentation, we use SAM with the pre-trained Vision Transformer Huge (ViT-h), the largest of three available Image encoders.
We then compare our method with ACE~\citep{ghorbani2019towards} and MCD~\citep{vielhaben2023multi}.

\noindent\textbf{Implementation Details.} We use SAM on 400 images per class, randomly sampled from the ImageNet1k training set, generating a comprehensive dataset of concept candidates. The generalization of the discovered concept is then verified using class images obtained from the ImageNet1k validation set. In the human-subject study, we implement MCD to ensure the discovery of at least five concepts for each class to ensure a fair comparison to the other methods.

\subsection{Understandability}
\label{subsec:understandability}

\begin{figure}[t] 
    \centering
    \includegraphics[width=0.46\textwidth]{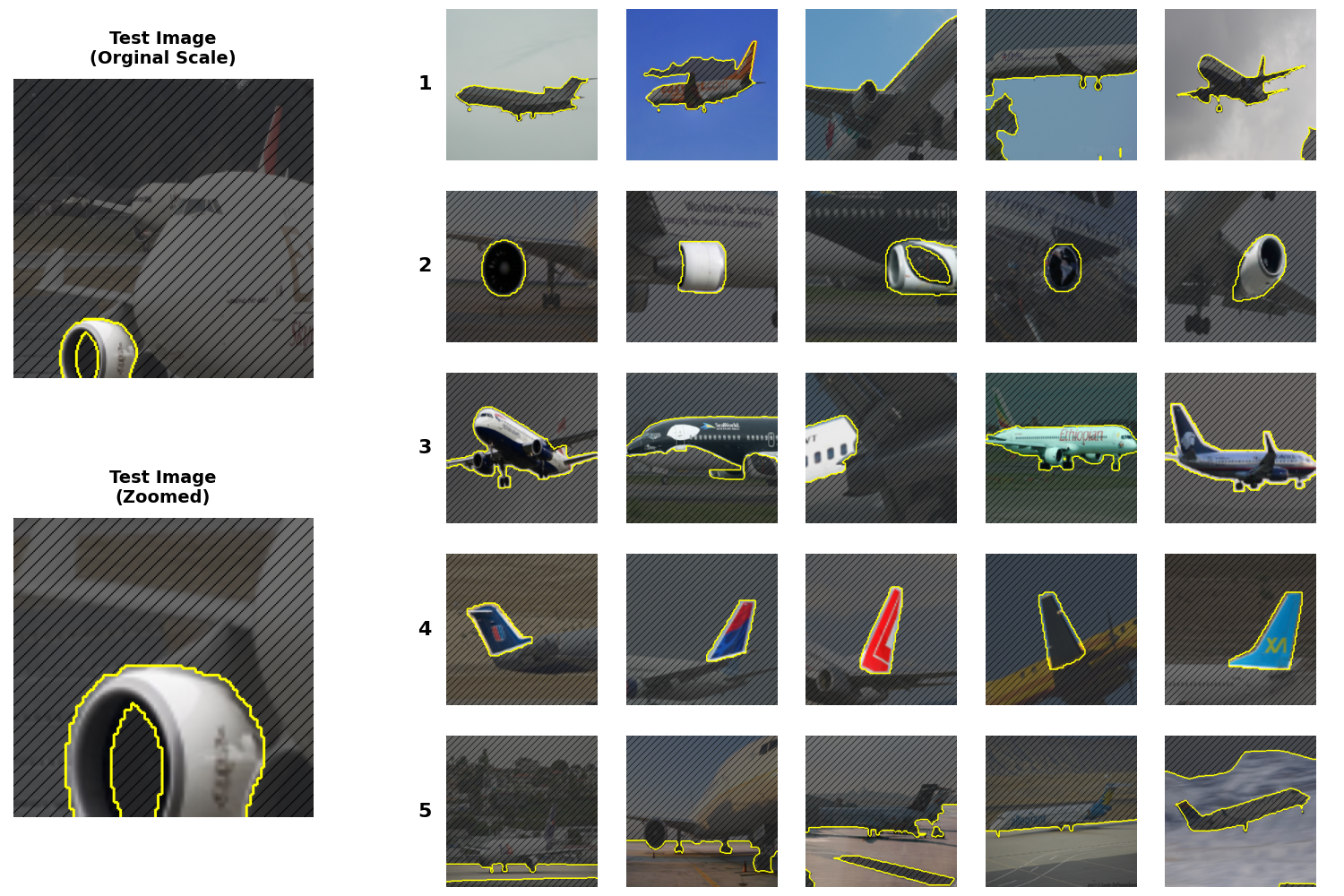}
    \caption{Survey sample of the human subject experiment generated by HU-MCD. Participants are asked to assign the test image on the left to the group on the right which is most similar by only considering the highlighted region.}
    \label{fig:survey_sample}
\end{figure}

\textbf{Human Subject Study Design.} Motivated by~\citet{zhang2021invertible}, we use task prediction~\citep{hoffman2023measures} for evaluation; participants are given one test image with one concept highlighted and five concept explanations from the same class as candidates (as shown in Fig.~\ref{fig:survey_sample}). They are then asked to select the candidate to which the test image most likely belongs. To account for ambiguity, participants can choose up to three or no candidates. Additionally, they are asked to rate the given concept candidates by stating whether they are \textit{recognizable} and assigning a 1-2 word description for each candidate they rate as recognizable. 

\noindent
The study used a within-subject design involving 15 examples from three methods (ACE, MCD, and HU-MCD) across five random CIFAR-10-like classes, presented in random order. Each participant encountered three sequential examples of each class, with method order randomized. Explanations were generated following method-specific details, retaining the top-10 most influential concepts for each class representing each concept by the top-10 prototypical samples. Five concepts were randomly selected as candidates for each test image. Ten random samples were created per class and method, for a total of 300 different samples. Each participant had a unique selection and order of samples and conducted a short tutorial.

\noindent
The experiment was conducted online using SoSci Survey, and participants were recruited through Prolific. Attention checks were included to ensure participant comprehension and exclude those who failed the checks. Participants with a prediction accuracy below 20\% (random choice) were excluded. A total of 41 participants completed the survey, and each experiment lasted approximately 30 minutes. Participants were compensated with £3 plus an incentive of 3 pennies per correct assignment, up to £3.45. Participant demographics were 61\% male, 37\% female, 2\% unspecified, with ages ranging from 18 to 68 ($\mu = 36$).

\begin{table}[t]
    \centering
    \begin{adjustbox}{width=.99\linewidth}
    \begin{tabular}{@{}cccccc@{}}
        \toprule
        & & \thead{\textbf{Prediction} \\ \textbf{Accuracy} $\uparrow$} & \thead{\textbf{Percentage} \\ \textbf{Recognizable} \\ \textbf{Concepts} $\uparrow$} & \thead{\textbf{Inner-Concept} \\ \textbf{Description} \\ \textbf{Similarity} $\uparrow$} & \thead{\textbf{Intra-Concept} \\ \textbf{Description} \\ \textbf{Similarity} $\downarrow$} \\
        \midrule
        \multirow{3}{*}{\makecell{\textbf{Results}}} 
            & HU-MCD & \textbf{70.24\%} & \textbf{67.12\%} & \textbf{0.49} & \textbf{0.28}  \\
            & ACE & 42.93\% & 45.66\% & 0.39 &  0.29   \\
            & MCD & 31.22\% & 50.34\% & 0.41 & 0.38  \\
        \midrule
        \multicolumn{2}{c}{\textbf{ANOVA test p-values}} & $<0.001$ & $<0.001$ & $<0.001$ & $<0.001$  \\
        \midrule
        \multirow{3}{*}{\makecell{\textbf{T-test} \\ \textbf{p-values}}} 
            & HU-MCD vs. ACE & $<0.001$ & $<0.001$ & $<0.001$ & 0.0155 \\
            & HU-MCD vs. MCD & $<0.001$ & 0.001 & 0.008 & $<0.001$ \\
            & ACE vs. MCD & 0.029 & 0.3773 & 0.4133 & $<0.001$ \\
        \bottomrule
    \end{tabular}
    \end{adjustbox}
    \caption{Results of the human-subject study to validate the \textit{understandability} of the concept explanations generated by HU-MCD. All results involving HU-MCD are statistically significant ($p<0.05$).}
    \label{tab:study_results}
\end{table}

\begin{figure*}[t!]
    \centering
    \includegraphics[trim=1.75cm 40cm 1cm 5cm, clip, width=\textwidth]{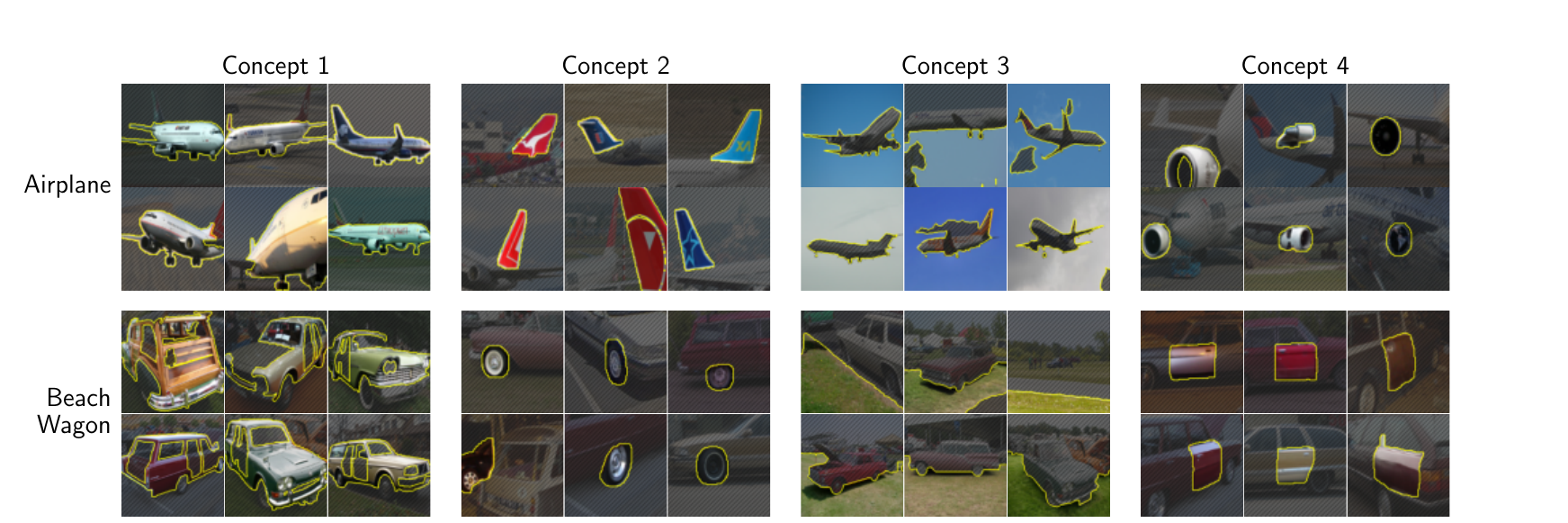}
    \includegraphics[trim=1.75cm 6cm 1cm 44.7cm, clip, width=\textwidth]{figures/case_study.pdf}
    \caption{Concept examples for three of the ten \textit{CIFAR-10} alike classes generated by HU-MCD.}
    \label{fig:concept_prototypes}
\end{figure*}

\noindent
\textbf{Metrics.} Similar to~\citet{zhang2021invertible}, we report the percentage of correctly identified concept explanations. The underlying assumption is that better concept explanations enable participants to associate the highlighted region within the test image to its corresponding concepts more accurately (among five candidates). High prediction accuracy indicates the coherence of individual concepts by requiring concept prototypes of the same concepts to be perceptually similar while being dissimilar to prototypes of other concepts. By asking participants to select all recognizable concepts, we further ensure the understandability of each generated concept explanation. Comparing the 1-2 word descriptions across participants serves as a supplementary indicator of the explanatory quality of the generated concepts. We use pre-trained GloVe~\citep{pennington2014glove} word vector representations for each description\footnote{We use the \textit{glove-wiki-gigaword-300} model loaded via the Gensim library~\citep{rehurek_lrec}}, and we then compute the average pairwise cosine similarity to assess the consistency of concept descriptions across participants (\ie, inner-concept description similarity). Understandable concept explanations should result in different participants assigning similar descriptions to the same concept.  Additionally, descriptions across different concepts should differ to indicate that the concepts characterize different attributes of a class. Thus, we also calculate the pairwise cosine similarity of the description embeddings across concepts within one class for each method (\ie, intra-concept description similarity).

\noindent
\textbf{Experimental Results.} As shown in Table~\ref{tab:study_results}, explanations generated by HU-MCD demonstrate a higher prediction accuracy over both ACE and MCD. This confirms that HU-MCD generates a diverse set of concepts and that prototypes of single concepts are perceived as perceptually similar. The superior understandability of the concept explanations is further supported by the percentage of concepts marked as recognizable. 
Interestingly, MCD shows a notable gap between prediction accuracy and the percentage of recognizable concepts. We observed that this is because the concepts generated by MCD for each class are highly similar, making it possible to recognize them individually but difficult to distinguish them from each other.
To underscore this, we additionally evaluated the percentage of recognizable concepts, considering only concepts for which participants provide unique descriptions within each question. This yields a proportion of 37,83\% (- 7.83\% in comparison to the values reported in Table~\ref{tab:study_results}) of \textit{uniquely} recognizable concepts for ACE, 38.43 \% (- 11.91\%) for MCD and 61.03\% (- 6.09\%) for HU-MCD. Notably, MCD shows the most substantial decline, indicating a lack of clear distinction among its discovered concepts, which limits their effectiveness in explaining the model's prediction behavior.

\noindent
Finally, HU-MCD achieves the highest description similarity for the same concepts across participants while maintaining distinct descriptions for different concepts within a class. This finding supports the understandability of HU-MCD concepts, as they are perceived similarly by multiple participants. In contrast, MCD shows minimal differences between intra- and inter-concept description similarity, suggesting that participants struggle to consistently identify the meaning of individual concepts and distinguish them. The reason for the improved distinguishability of concepts can be attributed to the usage of SAM, which is trained on human-annotated segmentation masks and thus produces interpretable segments.

\noindent
\textbf{Case Study.} Figure~\ref{fig:concept_prototypes} displays sample concepts identified by HU-MCD, represented by prototypical image segments from the ImageNet1k validation set. It clearly demonstrates the high quality of the segmentation masks, along with the perceptual similarity among prototypes representing the same concept and the distinctiveness between different concepts within the same class, each highlighting unique attributes of the class. These results align with findings from the human-subject study, confirming that HU-MCD's concept explanations are human-understandable.
HU-MCD identifies not only entire objects but also parts and contextual information, allowing humans to select and assess concept significance and completeness for any subset of discovered concepts. Notably, HU-MCD identifies concepts indicating spurious correlations among class images, such as \textit{human hands} (concept 4) for the ``\textit{tailed frog}.'' Quantifying the significance of such concepts in model predictions provides a valuable tool for systematic investigations into spurious correlations, thereby addressing model biases. 

\subsection{Faithfulness}
\label{subsec:faithfulness}

\noindent
\textbf{Metrics.} To validate the faithfulness of HU-MCD explanations and compare them to prior work, we use the Concept Deletion (C-Deletion) and Concept Insertion (C-Insertion) benchmarks~\citep{ghorbani2019towards}. C-Deletion identifies the smallest set of concepts whose removal results in an incorrect prediction, while C-Insertion identifies the smallest set sufficient for a target class prediction. For each sample, concepts are flipped (masked or unmasked) in decreasing order of local relevance, and the results are aggregated into a single line plot. Specifically, C-Deletion starts with the unmasked image and gradually masks concepts, while C-Insertion starts with the masked image and gradually reveals concepts. Faithful concept relevance scores result in a sharp decrease (C-Deletion) or increase (C-Insertion) in prediction accuracy with the number of flipped concepts.
We report the average model prediction accuracy as a function of the fraction of occluded pixels, as proposed by \citet{vielhaben2023multi}.

\noindent
\textbf{Experimental Setup.} To obtain concept masks and local importance scores for validation images, we segment each image using SAM, compute latent activation with input masking, and apply SSC. Each segment is assigned to a unique concept by selecting the maximum concept activation score and excluding segments aligned with the orthogonal complement, as these are assumed not to encapsulate substantial conceptual significance. The concept importance is calculated by averaging the local relevance score of the corresponding segments, resulting in ordered local concept masks. For ACE and MCD, we follow their original implementation details, noting that ACE uses TCAV scores instead of local concept importance scores. Results are then averaged over 500 images, using 50 ImageNet1k validation images for each of the ten CIFAR10-like classes, and only concepts present in at least 75\% of images across all classes are flipped to ensure meaningful averages.

\begin{figure}[t]
  \begin{minipage}[b]{0.49\linewidth}
  \centering
    \textbf{Concept Deletion}
    \includegraphics[width=\textwidth]{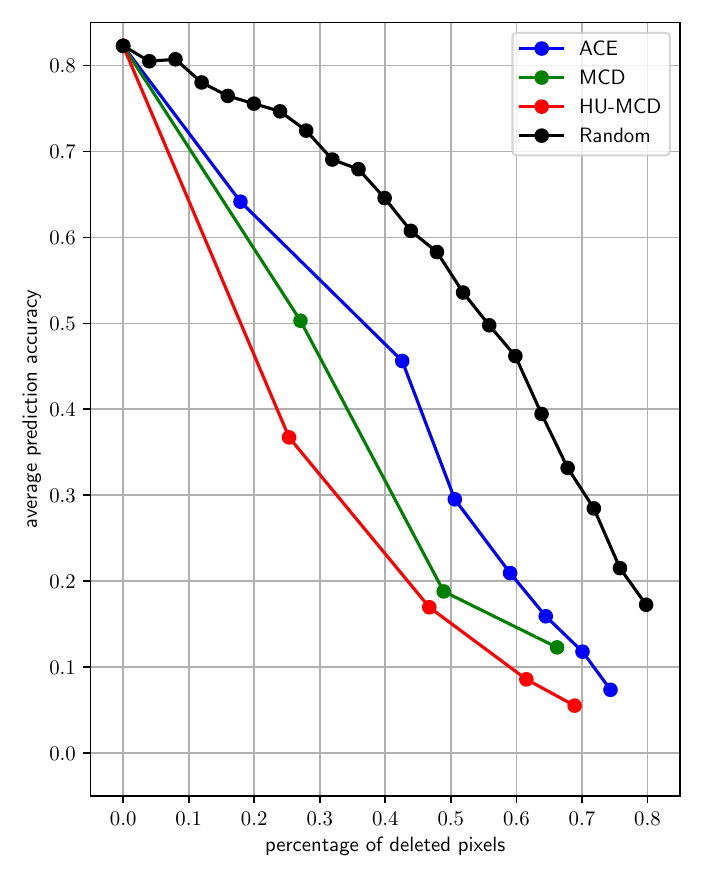}
  \end{minipage}
  \hfill
  \begin{minipage}[b]{0.49\linewidth}
    \centering
    \textbf{Concept Insertion}
    \includegraphics[width=\textwidth]{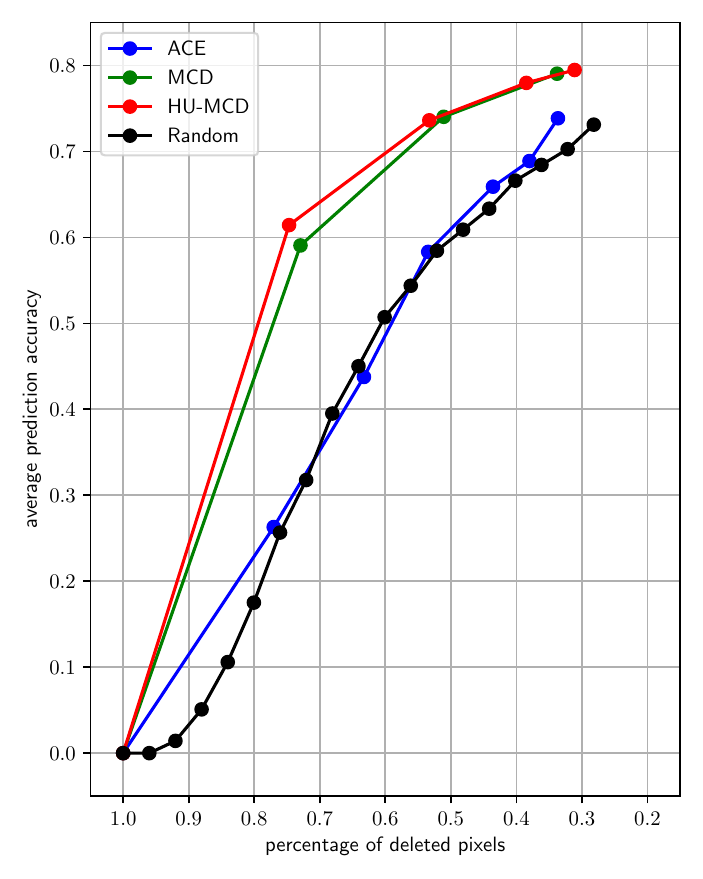}
  \end{minipage}
    \caption{We delete (left) or insert (right) concepts in decreasing order of concept importance and measure the impact on model prediction accuracy, averaged over all validation images of ten \textit{ImageNet1k} classes. Each point represents a discovered concept. Faithful concept importance scores are supposed to result in a sharp decline (left) or ascent (right).}
    \label{fig:concept_flipping}
\end{figure}

\noindent
\textbf{Experimental Results.} The results for both C-Deletion and C-Insertion are displayed in Fig.~\ref{fig:concept_flipping}. HU-MCD outperforms ACE and MCD in both benchmarks. This result emphasizes the faithfulness of the concept importance scores generated by HU-MCD in representing the model's reasoning process. Interestingly, the C-Insertion benchmark proves to be more challenging than C-Deletion, given that the model begins with a fully masked image, resulting in a near-random performance for ACE. Although the gap narrows, HU-MCD consistently achieves accuracy that is either superior to or comparable with MCD. 

\section{Limitations}

HU-MCD demonstrated promising results for the explanations' \textit{understandability} and \textit{faithfulness}. However, we acknowledge certain limitations. 
First, non-coherent concepts may arise due to segmentation errors, clustering inaccuracies, or limitations in the similarity metric. 
Such occurrences may be due to inherent methodological constraints or discrepancies between the model’s and humans’ perception of similarity. The user study shows that such occurrences are infrequent. Future work can entail an exploration of hyperparameters to overcome this challenge. 

\noindent
Second, our experiments are conducted on visual data, and the user study includes only ten classes. As emphasized by previous work~\citep{yeh2020completeness}, the general idea of concept-based explanations also applies to other data types, such as texts. 
This adoption challenge provides a promising avenue for future research to adapt HU-MCD to other modalities and expand its usability. Additionally, future research can extend the evaluation of HU-MCD to a wider variety as well as more fine-grained classes. 

\noindent
Finally, a technical limitation in the current setting is that the experiments are restricted to layers followed solely by linear operations to calculate concept relevance scores. However, concept activations can reveal the learned structures within the feature space for any layer.
As proposed by \citet{vielhaben2023multi}, future research could approximate the remainder of the model with a linear model, thereby enabling the quantification of concept relevance at different layers.
\section{Conclusion}

In this work, we introduced HU-MCD, a novel framework designed to extract human-understandable concepts from ML models automatically. HU-MCD satisfies two key criteria that previous research has neglected: (1) providing human-understandable concepts \textit{and} (2) faithfully attributing their importance to the model's predictions. For the first time, we use the SAM to extract understandable concepts automatically. Our approach extends prior concept discovery methods that use segmentation techniques by implementing a novel input masking scheme, which addresses noise introduced by inpainting and rescaling requirements. By representing concepts as multi-dimensional linear subspaces within the hidden feature space of a trained ML model, HU-MCD enables the decomposition of activations into unique concept attributions. This facilitates the calculation of concept importance scores both globally (per-class) and locally (per-image). Additionally, HU-MCD incorporates a completeness relation, quantifying the extent to which concepts sufficiently explain the model's predictions. This distinguishes it from most existing work in the field and offers the possibility to analyze models from a more human-centered perspective. We conduct extensive experiments on common benchmarks as well as a user study demonstrating both the faithfulness and understandability of the explanations generated by HU-MCD.
\newpage
\section*{Acknowledgment}
We thank Maria-Paola Forte from Max Planck Institute for Intelligent Systems for her valuable feedback on this work. We also thank the Karlsruhe Digital Service Research \& Innovation Hub (KSRI) team for the valuable feedback and consultation on the study design.

{
    \small
    \bibliographystyle{ieeenat_fullname}
    \bibliography{main}
}

\clearpage
\maketitlesupplementary

\section*{Evaluation Details}

\begin{figure}[h] 
    \centering
    \includegraphics[width=0.46\textwidth]{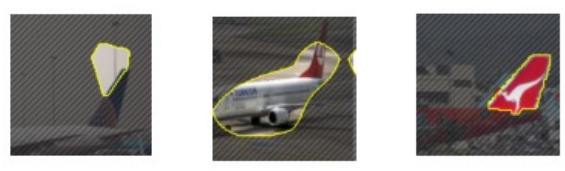}
    \caption{Example of outputs shown to participants: left ACE, middle MCD, right HU-MCD. Accuracy of descriptions can be found in Section 4.}
    \label{fig: exampleoutputs}
\end{figure}

In \Cref{fig: exampleoutputs}, we show examples that were shown to participants in the study. The figure shows an example of ACE (left), MCD (middle), and HU-MCD (right).

Beyond our primary evaluations, we report the number of discovered clusters and their completeness values for the ten CIFAR-10-like classes used in our experiments. The cluster count for each class is derived from the average number of image segments generated by SAM. To ensure concept coherence, we consider clusters with more than 50 segments as concepts, following \citet{ghorbani2019towards}. Using SAM's segmentation capabilities, we account for class-specific variability, recognizing that different classes exhibit different conceptual structures, resulting in a varying cluster count. We refer to \citet{vielhaben2023multi} for detailed algorithmic descriptions regarding clustering and experiments that verify these hyperparameter settings.

\begin{table}[h]
    \centering
{\small\fontsize{7}{12}\selectfont
    \resizebox{\linewidth}{!}{%
        \begin{tabular}{lcc}
            \hline
            Class & Number of Clusters & Completeness \\
            \hline
            Beach Wagon & 19 & 0.73 \\
            Hummingbird & 12 & 0.69 \\
            Police Van & 20 & 0.79 \\
            Ox & 15 & 0.67 \\
            Container Ship & 13 & 0.67 \\
            Siamese Cat & 15 & 0.75 \\
            Zebra & 12 & 0.69 \\
            Golden Retriever & 13 & 0.67 \\
            Tailed Frog & 14 & 0.68 \\
            Airliner & 11 & 0.65 \\
            \hline
        \end{tabular}%
    }
    \caption{Number of clusters and completeness scores of the discovered concepts for the CIFAR-10-like classes used in our experiments.}
    \label{tab:class_completeness}
    }
\end{table}

\section*{Effectiveness of the Input Masking Scheme}

In our experiments, we demonstrate the superior faithfulness of HU-MCD's generated explanations compared to ACE and MCD, which is particularly enforced through the application of the input masking scheme, aimed at ensuring that neither the baseline color used for inpainting the image segment nor the shape of the segmentation mask introduces undesired artifacts that could distort the model's prediction. 

\begin{figure}
    \centering
    \begin{subfigure}[t]{0.24\linewidth}
        \centering
        \includegraphics[width=\textwidth]{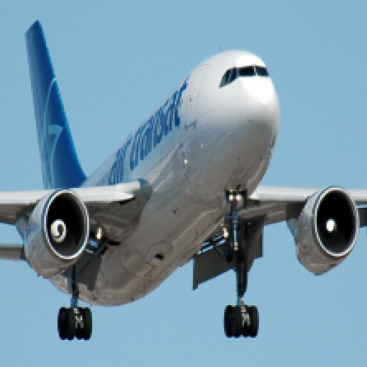}
        \captionsetup{justification=centering}
        \caption{Original Image}
    \end{subfigure}
    \hfill
    \begin{subfigure}[t]{0.24\linewidth}
        \centering
        \includegraphics[width=\textwidth]{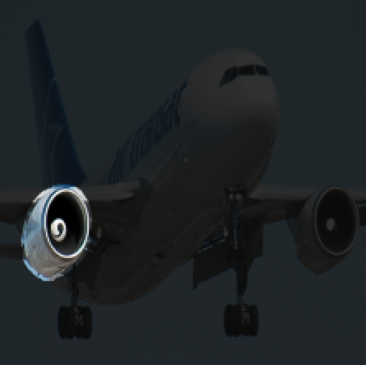}
        \captionsetup{justification=centering}
        \caption{Input Masking (original scale)}
    \end{subfigure}
    \hfill
    \begin{subfigure}[t]{0.24\linewidth}
        \centering
        \includegraphics[width=\textwidth]{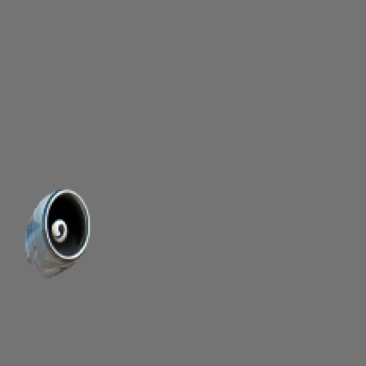}
        \captionsetup{justification=centering}
        \caption{Inpainting (original scale)}
    \end{subfigure}
    \hfill
    \begin{subfigure}[t]{0.24\linewidth}
        \centering
        \includegraphics[width=\textwidth]{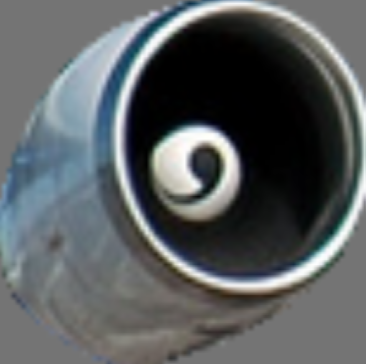}
        \captionsetup{justification=centering}
        \caption{Inpainting (cropped segment)}
    \end{subfigure}
    \caption{Different masking options. When employing Input Masking (option b), only the information within the highlighted region and its immediate surroundings is propagated through the model. In the case of inpainting (options c and d), the baseline color (grey) is propagated through the model.}
    \label{fig:masking_examples}
\end{figure}

\begin{figure}[t!]
  \begin{minipage}[b]{0.48\linewidth}
    \centering
    \textbf{Concept Deletion}
    \includegraphics[width=\textwidth]{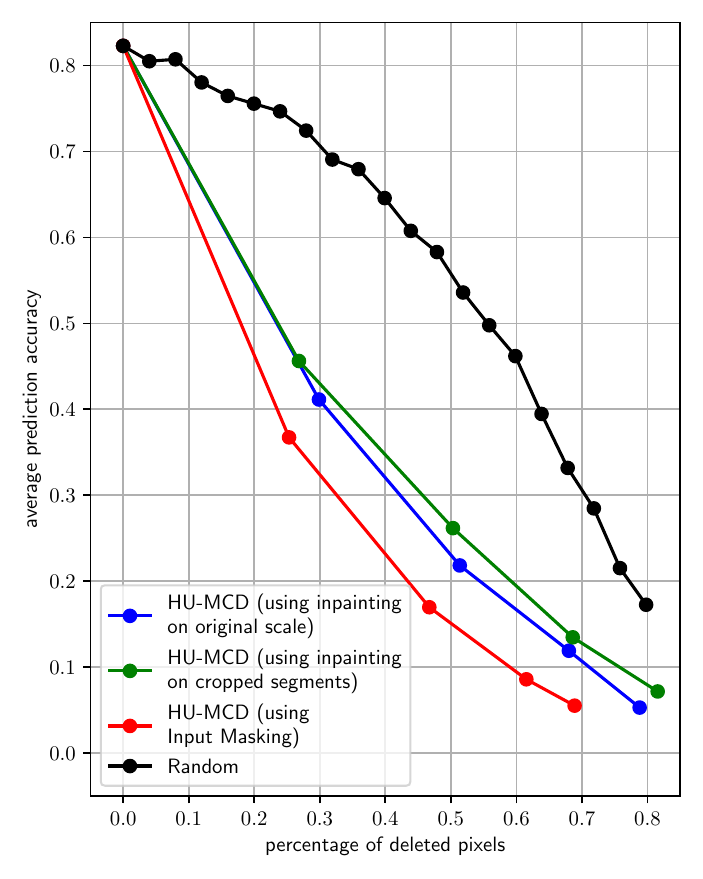}
  \end{minipage}%
  \begin{minipage}[b]{0.48\linewidth}
    \centering
    \textbf{Concept Insertion}
    \includegraphics[width=\textwidth]{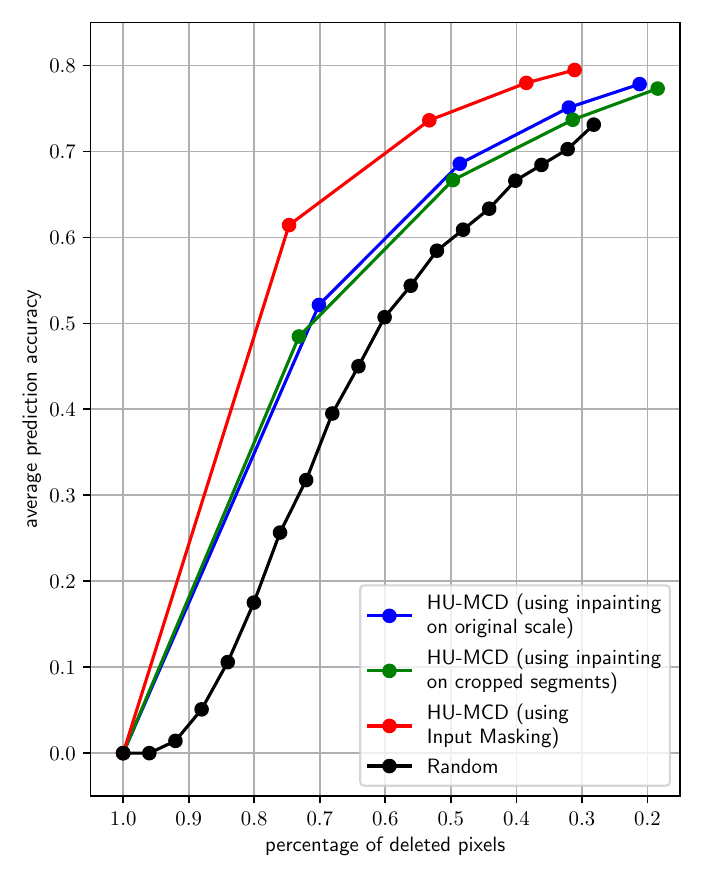}
  \end{minipage}
  \caption{We delete (left) or insert (right) concepts in decreasing order of concept importance and measure the impact on model prediction accuracy, averaged over all validation images of ten \textit{ImageNet1k} classes. Each point represents a discovered concept. Faithful concept importance scores are supposed to result in a sharp decline (left) or ascent (right). We compare the input masking scheme of HU-MCD to the simpler approach of masking with a baseline color, as used in previous methods, either at the original scale of the image or after cropping and resizing the image to fit the segment.}
  \label{fig:sdc_and_ssc_no_masking}
\end{figure}

We recalculate the C-Deletion and C-Insertion benchmarks for two alternative masking strategies: (1) masking regions outside segments with a baseline value and (2) cropping segments to minimal bounding boxes and rescaling (as in ACE). Figure~\ref{fig:masking_examples} shows the three different setups using an example image of class ``\textit{airliner}''. 

The results are illustrated in Figure~\ref{fig:sdc_and_ssc_no_masking}, which shows the effectiveness of the Input Masking scheme in ensuring that the concept importance scores \textit{faithfully} reflect the model's reasoning process. Interestingly, using inpainting at both the original scale and on the cropped segments yields similar performance, which is surprising, particularly considering that masks covering only small regions propagate much information regarding the baseline color through the network. Although CNNs are trained to be robust against scale variations, this result shows that scale information, as well as aspect ratio, plays a significant role, and distorting them negatively impacts the generated explanation quality.

\end{document}